\documentclass[runningheads]{llncs}

\usepackage{graphicx,wrapfig}
\usepackage{caption}
\usepackage{subcaption}
\usepackage{wrapfig}
\usepackage{geometry}
\usepackage{amsmath}
\usepackage{amsfonts}
\usepackage{mathtools}
\usepackage{hyperref}   
\hypersetup{
  colorlinks   = true,    
  urlcolor     = blue,    
  linkcolor    = blue,    
  citecolor    = red      
}
\usepackage{adjustbox}
\usepackage[dvipsnames]{xcolor}
\usepackage{authblk}
\usepackage{siunitx} 
\usepackage[subtle]{savetrees}
\usepackage{lipsum}

\begin{document}

\makeatletter
\newcommand\blfootnote[1]{%
  \begingroup
  \renewcommand{\@makefntext}[1]{\noindent\makebox[1.8em][r]#1}
  \renewcommand\thefootnote{}\footnote{#1}%
  \addtocounter{footnote}{-1}%
  \endgroup
}
\makeatother

\title{Understanding Tool Discovery and Tool Innovation Using Active Inference}

\author{%
Poppy Collis \thanks{Corresponding author} \inst{1}  \and 
Paul F Kinghorn \inst{1} \and
Christopher L Buckley \inst{1,2}
}%
\institute{
School of Engineering and Informatics, University of Sussex, Brighton, UK\\
\email{ \{pzc20, p.kinghorn, c.l.buckley\}@sussex.ac.uk},\\ 
\and
VERSES AI Research Lab, Los Angeles, California, USA\\}
\authorrunning{P. Collis, P. F. Kinghorn and C.L. Buckley}
\maketitle      

\blfootnote{P. Collis and P. F. Kinghorn—These authors contributed equally to this work.} 

\begin{abstract}
The ability to invent new tools has been identified as an important facet of our ability as a species to problem solve in dynamic and novel environments \cite{o2010innovation}. While the use of tools by artificial agents presents a challenging task and has been widely identified as a key goal in the field of autonomous robotics, far less research has tackled the invention of new tools by agents. In this paper, (1) we articulate the distinction between tool discovery and tool innovation by providing a minimal description of the two concepts under the formalism of active inference. We then (2) apply this description to construct a toy model of tool innovation by introducing the notion of tool affordances into the hidden states of the agent’s probabilistic generative model. This particular state factorisation facilitates the ability to not just discover tools but invent them through the offline induction of an appropriate tool property. We discuss the implications of these preliminary results and outline future directions of research.

    \end{abstract}

\section{Introduction}
Tool innovation has been identified as a core feature of human cognitive and cultural development, and has provided us with a key adaptive advantage as a species to survive adverse environments \cite{o2010innovation,stout2011stone,biro2013tool}. While both the use and innovation of tools was initially seen as a uniquely human capability, evidence has shown that a phylogenetically widespread variety of non-human animals engage in forms of tool manipulation, innovation and manufacture \cite{reader2016animal}. A large body of research has approached the topic of tool use in humans, animals and robotic systems \cite{Animaltoolusecurrentdefinitionsandanupdatedcomprehensivecatalog,cabrera2020neural,qin2022robot}. Here, we define tool use to be “the exertion of control over a freely manipulable external object (the tool) with the goal of altering the physical properties of another object, substance, surface or medium (the target, which may be the tool user or another organism) via a dynamic mechanical interaction” \cite{st2008revisiting}. However, developing the understanding of how to use a given tool is significantly different from the process of inventing a new tool.

Tool innovation refers to the process by which an agent independently constructs novel tools without relying on social demonstration or observation. This requires the ability to envision and conceptualise the appropriate tool for a given problem, while the knowledge of how to physically transform materials during construction is referred to as tool manufacture \cite{beck2011making}. The task of tool innovation presents a challenging problem in artificial agents, yet it is one that we as humans are inherently very good at. Indeed, research indicates that we develop innovation skills at a very early age \cite{breyel2021beginnings}. The animal innovation literature suggests that we can distinguish between two different classes of tool innovation: 1) that which arises as a result of incidental discovery where the animal then simply repeats this action in the same context and 2) that which is the result of intentional action by the animal resulting from some process of causal inference \cite{whiten2007evolution}. Herein, we define these two classes of innovation as \textit {tool discovery} and  \textit{tool innovation} respectively. 

Making such a distinction for both animals and human infants is challenging given the difficulty in determining the intentions driving subjects’ proposed solutions to a problem \cite{breyel2021beginnings}. While human behavioural experiments often explore the putative cognitive abilities required for tool innovation, no attempt is made to model such behaviour \cite{chappell2013development}. We therefore seek to offer a simple model of the cognitive phenomena underpinning the process of tool innovation. In the interest of a focused inquiry and to maintain conceptual clarity, we limit ourselves to being concerned with the causal reasoning involved in the process of tool innovation, while choosing to omit the challenges associated with the motor skills required to manipulate objects and manufacture tools from physical materials.

In recent years, theories which describe the brain as broadly Bayesian have gained considerable traction in the field of neuroscience. The ‘Bayesian brain hypothesis’ posits that perception arises as a result of Bayesian model inversion, with incoming sensory data updating these causal models of the world in accordance with Bayes’ rule \cite{friston2012history}. The theory of active inference (AIF) extends this idea and casts action, perception and learning as being underwritten by the same underlying process of Bayesian inference. Derived from first principles, the theory provides a formal account of behaviour arising as a result of the imperative to minimise of the information-theoretic quantity of surprisal. In other words, an autonomous agent is continually in the act of accumulating Bayesian model evidence (``self-evidencing'') and it is from this perspective that we can understand decision-making under uncertainty \cite{sajid2021active}. AIF offers a rich description of the internal mechanisms of belief-based reasoning and principled account of the natural emergence of curious and insightful adaptive behaviour \cite{friston2017active}. It has also recently been proposed as a framework well-suited to robotics \cite{da2022active}. We therefore chose to explore the concept of tool innovation using this framework.

The main contribution of this paper is (1) the articulation of the distinction between tool discovery and tool innovation within the AIF framework and (2) a minimal model of non-trivial tool innovation that requires generalised inferences about the tool structure required to solve a task. First of all, we show that with a perfect generative model, the agent can straightforwardly use tools optimally to solve a task. We then demonstrate that the agent can discover the correct tools and learn to solve the task when it is not provided with this information in its model \textit{a priori}. Finally, we provide evidence that factorising the hidden states of the generative model into the affordances of the tool can enable the agent to conceive offline the appropriate properties of the tool required to solve the task. It is this difference between the generative model and the generative process which is key to facilitating tool invention. This enables the agent to not simply happen upon the appropriate tool during exploration of environmental contingencies, but to invent them through the induction of an appropriate tool property. We discuss the implications of these preliminary results and outline future directions of research.

\section{Active Inference in Discrete State Space}

In AIF, the minimisation of sensory surprisal is achieved through the minimisation of a tractable quantity called the variational free energy $\mathcal{F}$, known as (negative) evidence lower-bound (ELBO) in the variational inference literature \cite{blei2017variational}. This minimisation is performed via the maintenance of a probabilistic generative model of the environment. AIF has been widely implemented using discrete-time stochastic control processes known as partially-observable Markov decision processes (POMDPs) \cite{da2022active}. We therefore implement our simulations agent with an AIF framework in discrete state space using the Python package \textit{pymdp} \cite{heins2022pymdp}. This specifies a standard POMDP generative model as a joint probability distribution over observations $o$, hidden states $s$, policies $\pi$ and model parameters $\phi$. In contrast to much of the reinforcement learning literature, a policy in this case is defined as a fixed sequence of control states $u_\tau$ for each timestep $\tau$ that together represent a plan of action of length $T$, $\pi = \{u_1, \dots ,u_T\}$ \cite{sajid2021active}. We assume the standard factorisation of the POMDP as a product of conditional (likelihood) distributions and prior distributions over a finite time horizon $[1:T]$.

The most important distributions when specifying this generative model are the observation likelihood $P(o_\tau \mid s_\tau ; \phi)$, the transition likelihood $P(s_\tau \mid s_{\tau-1}, \pi ; \phi)$, and the prior preference over observations $P(o_\tau)$, known in \textit{pymdp} as the A, B and C matrices respectively. We also further factorise our representations of $o_\tau$ and $s_\tau$ into separate modalities and factors: $o_\tau = \{o_\tau^1, o_\tau^2, \dots , o_\tau^M \}$ and $s_\tau = \{s_\tau^1, s_\tau^2, \dots , s_\tau^F \}$ in which $M$ is the number of modalities and $F$ the number of hidden state factors such that the likelihood distributions can be written as:
\begin{equation}
\label{obs_likelihood}
P(\mathbf{o_\tau} \mid \mathbf{s_\tau}) = \prod_{m=1}^M P(o_\tau^m \mid \mathbf{s_\tau})
\end{equation}

\begin{equation}
\label{state_transition}
P(\mathbf{s_\tau} \mid \mathbf{s_{\tau-1}}, \mathbf{u_{\tau-1})} = \prod_{f=1}^F P(s_\tau^f \mid \mathbf{s_{\tau-1}}, \mathbf{u_{\tau-1})}
\end{equation}

We allow state factors in the transition likelihoods to depend on themselves and a specified subset of other state factors.\footnote{This requires a recent branch of \textit{pymdp} which enables this kind of factorisation. See \url{https://github.com/infer-actively/pymdp/tree/sparse_likelihoods_111}} Since we are working in discrete space, the probability of states and observations can be described by a categorical probability distribution.

\begin{figure}[t]
    \includegraphics[width=6cm]{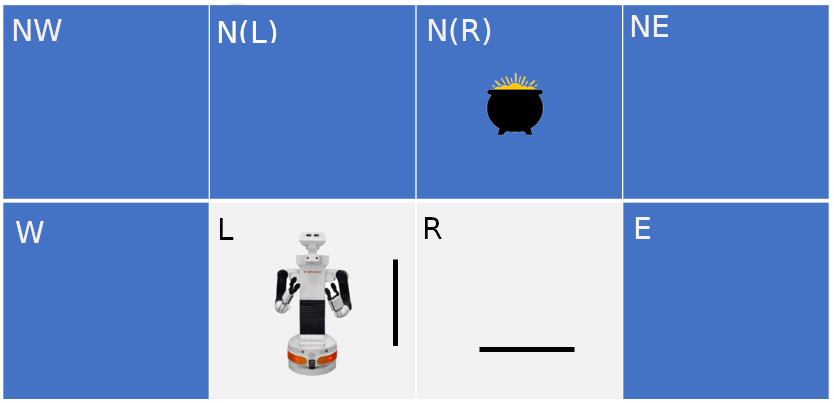}
    \centering
\caption{The task of the the agent is to reach the reward by using the  tools provided. The simulation environment shows the agent (\textit{robot}) can only move between the left and right rooms (\textit{grey}) and the reward (shown as a pot of gold) can be placed in any of the other rooms (\textit{blue}). A vertical tool (V) can be picked up in the left room, and a horizontal tool (H) in the right room}
\label{environ}
\end{figure}

In this work, we consider the simple environment shown in Fig.~\ref{environ}. It consists of a 2 x 4 grid of locations in which the agent can only move between two rooms: left and right (shown here in grey). The agent is always initialised in the left-hand room. A vertical tool (V) is located in the left-hand room while a horizontal tool (H) is located in the right-hand room. In one of the remaining rooms (shown in blue), a reward is located, and it is the goal of the agent to try and reach this reward using the tools provided. For example, if the reward is in the room directly north of the right-hand room as shown, the agent is required to be in the right-hand room holding tool V in order to reach it. The agent can choose to pick up the tool if it is in the relevant room, while it may drop tools whilst it is in any room (in which case, any of the tools in the agent's possession are dropped and returned to their original rooms). If the agent already possesses a tool and picks up a different tool, this creates a compound tool (HV). The rooms directly north, east and west of the left and right rooms are known as the \textit {adjacent rooms} and these only require the individual tools V or H to solve. The northeastern and northwestern rooms are termed the \textit {corner rooms} and present a greater challenge for the agent as they require the construction of the compound tool (HV) to solve.

\begin{table}
\centering 
\captionsetup{font=small}
\caption{Generative Model Structure} 
\begin{tabular}{l c c rrrrrrr} 
\hline\hline 
 States & Factors & Dimensions & Values
\\ 
\hline 
Hidden states & Room & 2 & Left, Right\\
 & Tool & 4 & Null, V, H, HV\\
\hline 
Observations & Room & 2 & Left, Right\\
 & Tool & 4 & Null, V, H, HV\\
 & Reward & 2 & Null, Reward\\
\hline 
Control States &  &  4 & Null, Move, Pick-up, Drop\\
\hline 
\end{tabular}
\label{tab:genmodel}
\end{table}

For the initial experiments, the hidden states of the environment are factorised into two factors, $s_\tau=\{s_{\tau}^1,s_{\tau}^2\}$, which consist of: room state and tool state (see Table.~\ref{tab:genmodel}). A policy length of 4 time-steps is chosen given that the task of retrieving the reward can always be solved optimally within 4 steps (for any reward location). As we have set the policy length to be 4 time-steps and we have 4 possible actions, we therefore have 256 ($4^4$) possible policies which we must individually evaluate by calculating the expected free energy for every time-step (see Section \ref{polinf}). In all experiments, the agent is equipped with a strong prior preference for the observation of reward in the reward modality. In terms of relative log probabilities, we specify this to be 0 for an observation of null and 50 for an observation of reward. Observations in all other modalities have a flat prior (i.e. no preference given).

\section{Policy Inference}
\label{polinf}
In AIF, policy inference is effectively a search procedure in which a free energy functional of expected states and observations under a policy is evaluated for each possible policy. Once we have calculated this quantity (known as the expected free energy, $\mathcal{G}$) for each policy, we can convert this into a probability distribution over the set. Action selection then simply amounts to sampling from this distribution accordingly. Policies which most minimise $\mathcal{G}$ will be assigned a higher probability and are therefore more likely to be chosen. Since the variational posterior factorises over time, we can calculate $\mathcal{G}$ for each time step independently. The expected free energy for a particular future time step under a particular policy is given by:

\begin{equation}
\label{G_original}
\mathcal{G}_\tau(\pi) = \mathbb{E}_Q[\ln Q(s_\tau|\pi) - \ln \tilde{P}(o_\tau , s_\tau|\pi)]
\end{equation}

where $\tilde{P}(o_\tau , s_\tau|\pi) = P(s_\tau| o_\tau, \pi)\tilde{P}(o_\tau)$, representing a generative model that is biased to produce preferred observations (for full derivations, see \cite{heins2022pymdp}). $\mathcal{G}_\tau(\pi)$ can be rearranged in various ways to give intuition about what it actually represents. One such representation decomposes this free energy functional into an epistemic value (information gain) term and a pragmatic value (utility) term:

\begin{equation}
\label{G_decomposed}
\mathcal{G}_\tau(\pi)\le\underbrace{-\mathbb{E}_{Q(o_\tau|\pi)}[D_{KL} \left[ Q(s_\tau|o_\tau,\pi)\parallel Q(s_\tau|\pi)\right]]}_\text{State Information Gain}-\underbrace{\mathbb{E}_{Q(o_\tau|\pi)}[\ln \tilde{P}(o_\tau)]}_\text{Utility}
\end{equation}

Epistemic value refers to the information gain from the expected outcomes of hidden states. Given a policy, it measures the divergence between the expected states and the expected states conditioned on the observations. This gives rise to curious behaviour in which the agent is compelled to minimise uncertainty about its environment via exploration. On the other hand, the utility term simply measures the extent to which the observations expected under a policy align with the observations the agent wishes to encounter. This promotes the exploitation of knowledge in order to satisfy preference over outcomes. This trade-off between exploration and exploitation therefore naturally arises in AIF; both imperatives are cast as ways in which an agent acts to resolve uncertainty.

\begin{figure}
 \vspace{-0.4cm}
 \captionsetup{justification=raggedright}
\centering
\begin{subfigure}{.5\textwidth}
  \centering
  \includegraphics[width=.8\linewidth]{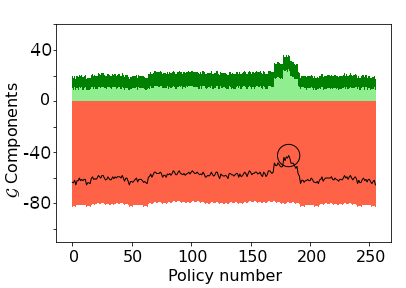}
    \caption{Information gain dominating policy selection}
  \label{policies:1}
\end{subfigure}%
\begin{subfigure}{.5\textwidth}
  \centering
  \includegraphics[width=.8\linewidth]{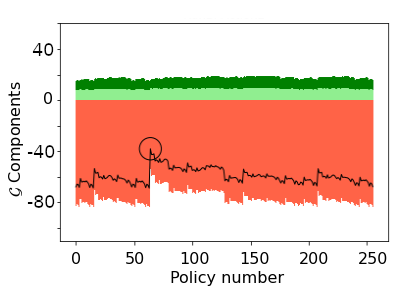}
  \caption{Utility dominating policy selection}
  \label{policies:2}
\end{subfigure}

\caption{Decomposing expected free energy $\mathcal{G}$ into respective information gain and utility contributions can elucidate the agent’s intended consequences of an action. The expected free energy $\mathcal{G}$ (\textit{black line}) is evaluated over a set of 256 policies. The components which contribute to the selection of the best policy (\textit{circled}) are state information gain (\textit{dark green}), parameter information gain (\textit{light green}) and utility (\textit{orange}). Examples shown are instances when the selected policy is a) driven by information gain as there is little variation in utility and b) driven by utility as there is little variation in information gain} 
\label{policies}
\vspace{-0.4cm}
\end{figure}

We can visualise this trade-off by plotting the respective utility and information gain components of the total $\mathcal{G}$. Fig.~\ref{policies:1} shows an example in which each of the policies vary little with respect to their expected utility and the policy selected has been driven by the high information gain component. In contrast, Fig.~\ref{policies:2} shows an example of when the dominant driving force in policy selection is the utility component while information gain remains largely invariant across policies. Note that we also include a parameter information gain term which is explained in Section \ref{sec4}.

\section{Parameter Inference}
\label{sec4}
Learning in AIF is a process of inference over the model parameters, $\phi$, which are simply the categorical likelihood distributions. We treat these parameters as something over which the agent maintains and updates beliefs (i.e. as random variables). Consider the example of an A matrix, which encodes the observation likelihood model $P (o | s )$, with the entry $A[i,j]$ representing the probability of seeing observation $i$ given state $j$. There is therefore a separate categorical distribution for each state (i.e. each column sums to 1). The Dirichlet distribution is a conjugate prior for the categorical distribution, and we therefore model prior beliefs over the categorical as a Dirichlet. It can be shown that, when the agent obtains new empirical information, the Bayesian process of updating this prior is simply a count-based increase of the Dirichlet parameters according to the observation $o$ and inferred state $s$ \cite{heins2022pymdp,murphy2012machine}:

\begin{equation}
\label{dirichlet}
\alpha_{posterior} = \alpha_{prior} +o\otimes s
\end{equation}

where $\alpha$ represents the Dirichlet parameters. Now that we are treating model parameters as random variables, we can expand $\mathcal{G}$ to include the expected parameter information gain component:

\begin{equation}
\label{G_decomposed_fully}
\mathcal{G}_\tau(\pi)\le\underbrace{-\mathbb{E}_{Q(o_\tau|\pi)}[D_{KL} \left[ Q(s_\tau|o_\tau,\pi)\parallel Q(s_\tau|\pi)\right]]}_\text{State Information Gain}
\underbrace{-\mathbb{E}_{Q(o_\tau|\pi)}[D_{KL} \left[ Q(\phi|o_\tau,\pi)\parallel Q(\phi|\pi)\right]]}_\text{Parameter Information Gain}-\underbrace{\mathbb{E}_{Q(o_\tau|\pi)}[\ln \tilde{P}(o_\tau)]}_\text{Utility}
\end{equation}
This will drive the agent to seek observations which lead to a larger change in the categorical distribution.

\section{Experiment 1: Tool Use}
\label{tooluseexp}
In the first set of experiments, the agent has a perfect probabilistic generative model of the world. This means that the correct transition likelihood and observation likelihood distributions are provided and therefore no learning is required. We then show that the agent can straightforwardly infer the optimal actions in order to achieve its goal of reaching the reward. We use this as a simple model of tool use in an autonomous agent, given the definition of tool use defined previously \cite{st2008revisiting}. By this account, our simulated agent conducts tool use by acting to “exert control over” tools V, H or HV in order to “alter the physical properties” of the tool user (by extending the agent’s reach) enabling it to successfully retrieve the reward. In this sense, we reduce tool use to an action sequencing problem.

\begin{table}
\centering 
\captionsetup{font=small}
\caption{Comparing optimal number of steps
required to solve each reward location with actions taken. When the generative model is perfectly known, the agent solves the task optimally\\} 
\begin{tabular}{l c c rrrrrrr} 
\hline\hline 
 Reward Location & \; Optimal No. of Steps & Actions of Agent\\ 
\hline 
North-left & 1 & Pick-up\\

North-right & 2 & Pick-up, Move\\

East & 2 & Move, Pick-up\\ 

West & 3 & Move, Pick-up, Move\\

Northeast & 3 & Pick-up, Move, Pick-up\\

Northwest & 4 & Pick-up, Move, Pick-up, Move /\\
& & Move, Pick-up, Move, Pick-up\\
\hline 
\end{tabular}
\label{tab:optimal}
\end{table}

For each trial, we place the reward in one of the possible reward locations and allow the agent 12 time-steps in which to act in the world and obtain the reward. The agent uses all 12 time-steps, and therefore if it has found the reward, the optimal behaviour would be to perform an action that will keep it in the same state (i.e. the action ``Null'').

\begin{figure}
    \includegraphics[width=7cm]{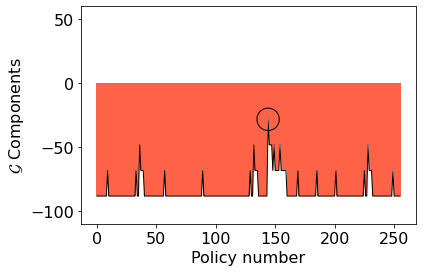}
    \centering
\caption{When the generative model is perfectly known, the selected policy is based solely on the utility component of $\mathcal{G}$. An example of $\mathcal{G}$ (\textit{black line}) evaluated for all 256 policies and the selected policy (circled) which is the one with the highest utility (\textit{orange}). Note that since the generative model is fully known, and the environment is fully observed, all policies have zero information gain component} 
\label{expt1}
\end{figure}

As expected, the agent solves the task of obtaining the reward optimally for each reward location (Table.~\ref{tab:optimal}). Given the stochastic nature of policy selection, we note in  that the agent solves the northwest room via two different methods, yet both are optimal (i.e. of length 4). Since the generative model is fully known, the agent gains no new information about states or parameters during inference. Indeed, Fig.~\ref{expt1} shows that if we plot the relative utility and information gain contributions to the expected free energy of each policy during action selection, we see that it only comprises of a utility component compared to Fig.~\ref{policies} (i.e. there is no epistemic value contribution to $\mathcal{G}$).

\section{Experiment 2: Tool Discovery}

Next, we investigate the ability of the agent to learn how a particular tool solves the task. We present this as a toy example of tool discovery given that knowledge about how to create a tool arises incidentally as a result of environmental exploration. Whilst we again provide the agent with a fully known observation likelihood distribution, for the following experiment we initialise the agent with a uniformly distributed transition likelihood model. This means that the agent initially knows nothing about how states and actions effect future states. It therefore must learn these state transitions rather than being provided with this information from the outset (as in experiment 1). The agent happens upon the correct tool to use for a given reward location, and then repeats this action in the same contexts. This is in line with our previous definition of tool discovery \cite{whiten2007evolution}.

\begin{figure}
 \vspace{-0.4cm}
 \captionsetup{justification=raggedright}
\centering
\begin{subfigure}{.5\textwidth}
  \centering
  \includegraphics[width=.8\linewidth]{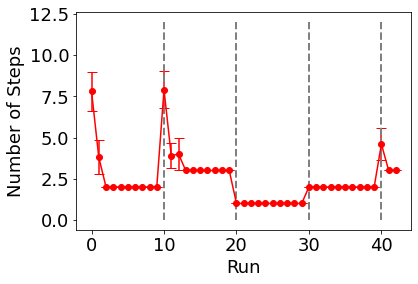}
    \caption{Finishing with reward in northeast room}
  \label{expt2:rm2}
\end{subfigure}%
\begin{subfigure}{.5\textwidth}
  \centering
  \includegraphics[width=.8\linewidth]{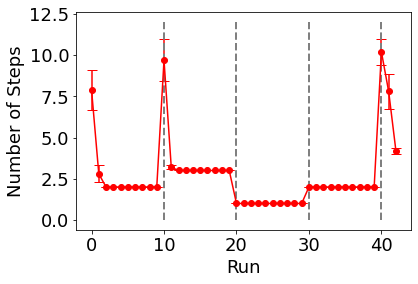}
    \caption{Finishing with reward in northwest room}
  \label{expt2:rm5}
\end{subfigure}
\caption{The number of steps taken to solve the task for each reward location decreases quickly over runs to the optimal number of steps, reflecting the agent learning via discovery. Graphs show the mean (+/- ste) number of steps to solve reward location averaged over 20 independent trials. The agent is exposed to a different reward locations every 10 runs (\textit{dashed lines}). The reward is first located in the adjacent rooms (in the order north-right, west, north-left, east) before being presented with a) the northeast or b) the northwest room for final 3 runs (40-42). For both cases, despite learning how to create a V and H tool in the earlier runs, the agent still has to learn about the HV tool when the reward is placed in a corner room} 
\label{expt2}
\vspace{-0.4cm}
\end{figure}

Fig.~\ref{expt2} shows that the number of steps the agent takes to find the reward decreases over the number of runs. In this continual learning task, each time the reward location changes (at runs 0, 10, 20, 30 and 40) it demands the learning of a new tool and we see an initial increase in the number of steps requires to solve the task. This is because the information the agent has about state transitions (i.e. how states and actions give rise to states at the next time-step) is not sufficient to solve the task. The agent therefore explores more of the environment before encountering the correct tool required to satisfy its preference for the reward observation. We then see a sharp drop after the agent has learned about the required state transitions, and the number of steps taken to solve the task quickly plateaus to the optimal number shown in Table.~\ref{tab:optimal}. 

Interestingly, as a result of the ordering in which the reward locations are presented (north-right, west, north-left, east, ...), the agent solves the north-left and east reward locations optimally from the outset. This is due to the fact that the solving of previous adjacent rooms (north-right and west) resulted from the learning of tool V and H respectively. When the agent then encounters the reward in the remaining adjacent rooms, it has already learned about the correct actions to create these tools to solve the task despite never having seen these particular reward locations before. The corner rooms require more steps despite having already learned V and H, as the agent must still discover the new tool HV. Given that the agent is always initialised in the left-hand room, the northwest corner (Fig.~\ref{expt2:rm5}) takes more steps to solve that the northeast corner (Fig.~\ref{expt2:rm5}) because it involves a more complex action sequence to retrieve the reward (see Table.~\ref{tab:optimal}).

\begin{figure}
 \vspace{-0.4cm}
 \captionsetup{justification=raggedright}
\centering
\begin{subfigure}{.33\textwidth}
  \centering
  \includegraphics[width=.9\linewidth]{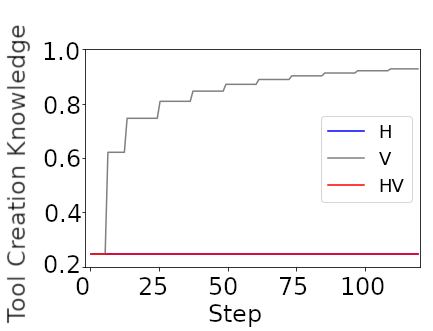}
    \caption{North-right room}
  \label{learning:rm7}
\end{subfigure}%
\begin{subfigure}{.33\textwidth}
  \centering
  \includegraphics[width=.9\linewidth]{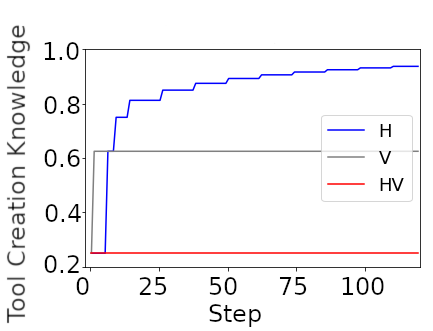}
    \caption{East room}
  \label{learning:rm3}
\end{subfigure}
\begin{subfigure}{.33\textwidth}
  \centering
  \includegraphics[width=.9\linewidth]{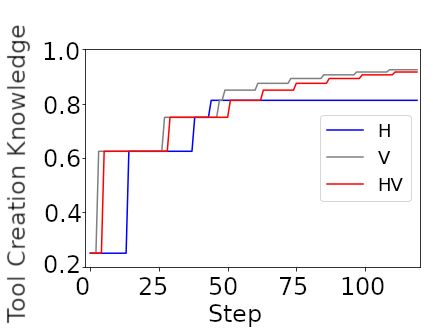}
    \caption{Northeast room}
  \label{learning:rm2}
\end{subfigure}
\caption{The agent only learns the tools that it needs to learn in order to solve the task. We provide a measure of how well the agent knows each tool by looking at the posterior probability associated with the correct control state (i.e. action) for creating each tool when solving for rooms a) north-right b) east and c) northeast over 125 steps} 
\label{learning}
\vspace{-0.4cm}
\end{figure}

Importantly, given that the minimisation of $\mathcal{G}$ naturally incorporates two competing imperatives (utility and information gain), this means that the agent learns only the tools that it needs to learn in order to solve the task, and does not continue exploring its environment if it is able to leverage its current knowledge to effectively realise prior preferences. Fig.~\ref{learning:rm7} shows that for the north-right room, the agent only learns the vertical tool (V). This is because the first tool it picked up (V) allowed it to solve the task and therefore the agent did not need to continue exploring the hidden states of the environment as it had all of the knowledge it needed. Fig.~\ref{learning:rm3}) shows that for the east room, the agent first tried the vertical tool (V), however this did not lead to the agent observing preferred observations (reward) and therefore it does not infer the action of picking up this tool again. Instead, it pursues policies which yield high information gain (i.e. it explores new states of the environment) and finds that picking up tool H leads to a rewarding observation. By selecting policies which maximise utility, it therefore repeats this action (``pick-up'') in the same context, and learns this tool with more confidence while neglecting to explore other options. Finally, Fig.~\ref{learning:rm2} shows that in order to discover the compound tool (HV), the agent first happens upon tools V and H (as these tools are more likely to be stumbled across given they require a less complex sequence of actions in order to learn about them). However, these do not provide it with high utility. Since there are unknown states (such as tool HV) that provide it with high information gain, the agent continues exploring and then finds that creating the compound tool brings about its preferred observations.

\begin{figure}
 \vspace{-0.4cm}
 \captionsetup{justification=raggedright}
\centering
\begin{subfigure}{.5\textwidth}
  \centering
  \includegraphics[width=.8\linewidth]{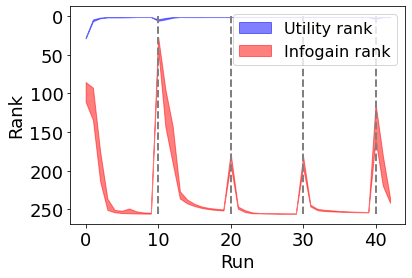}
    \caption{Finishing with reward in northeast room}
  \label{ranks:rm2}
\end{subfigure}%
\begin{subfigure}{.5\textwidth}
  \centering
  \includegraphics[width=.8\linewidth]{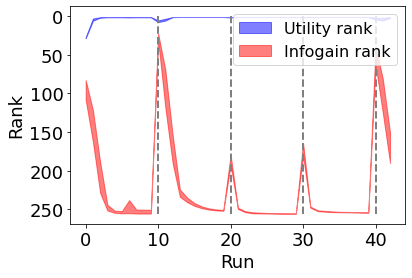}
    \caption{Finishing with reward in northwest room}
  \label{ranks:rm5}
\end{subfigure}
\caption{ Policy selection is initially dominated by information gain, but is then very quickly driven by utility as the agent learns new information. Graph shows how the selected policy ranks in the context of all possible policies in terms of utility and information gain (averaged over 20 independent trials) (best rank is 0, worst is 256). Like Fig.~\ref{expt2}, the reward location changes every 10 runs (\textit{dashed lines}) in the order north-right, west, north-left, east. The agent is then presented with a) the northeast or b) northwest room for the final 3 runs} 
\label{ranks}
\vspace{-0.4cm}
\end{figure}

A policy is selected on its value of $\mathcal{G}$ which is composed of both expected utility and expected information gain. We can visualise the evolution of this trade-off in driving policy selection during a continual learning trial. For each time-step, we see how the chosen policy ranks in the ordered list of all policies with respect to utility and the ordered list of all policies with respect to information gain. This rank provides us with a measure of the relative contributions of utility and information gain in the selection of a policy. For example, if the chosen policy ranks very highly for utility, and yet ranks very low in the context of the best policies for information gain, we know that the policy (and therefore resultant action) has been selected primarily due to its high utility.

As Fig.~\ref{ranks} shows, for each reward location, the information gain component is initially very high and therefore dominates action selection. This is because when the reward location is changed, the state transition information is not adequate to solve the task. The gain in information quickly drops as the agent learns transitions via exploration, while the utility rank of the policy increases as it can leverage this newly learned information to seek the preferred observation of the reward. Note that at runs 20 and 30, this spike in information gain is lower that at 0 and 10. This is because the agent has already learned about creating tool V and H in the north-right and west reward locations respectively. When the agent is then presented with the novel adjacent reward locations (north-left and east), it has the advantage of already having the knowledge of how to pick up the correct tool to use to solve the problem. For the final reward location (northeast for Fig.~\ref{ranks:rm2} and northwest for Fig.~\ref{ranks:rm5}), we also see a spike in information gain. This is in agreement with Fig.~\ref{expt2} which shows that we do indeed see an increase in the number of steps taken to solve these final rooms. Despite having knowledge about the individual tools H and V, the agent must explore further to ‘discover’ the compound tool.

We have therefore shown that the agent can leverage the knowledge gained in the incidental discovery of required state transitions to solve the task. This amounts to a simple model of tool discovery behaviour in accordance with our previously defined definition.

\section{Experiment 3: Tool Innovation}

The following experiment investigates the concept of tool innovation in our AIF agent. In order to achieve this, the agent must be able to analyse the problem and identify the kind of the tool required to solve the task. This entails developing a grounded understanding of the objects in the environment which can then be leveraged to construct a suitable tool through a process of generalisation. For the acquisition of grounded knowledge about the world, we turn to the concept of ‘affordances’ from ecological psychology \cite{gibson1977theory}. This refers to opportunities for action provided by the environment. In the robotics literature, this is defined as the “relationship between an actor (i.e., robot), an action performed by the actor, an object on which this action is performed, and the observed effect” \cite{andries2018affordance}. 

We adjust our generative model to incorporate the following tool affordances into the hidden states: the horizontal reach (x-reach) and vertical reach (y-reach) afforded by each tool and the room state $s_\tau=\{s_{\tau}^1,s_{\tau}^2,s_{\tau}^3\}$. Each affordance state can take a binary value. We refer to this as the \textit{Affordance Model} while the previous model which included an unfactorised tool state is referred to as the \textit{Tool State Model}. Importantly, these affordances do not depend on one another, which allows for generalisation of learning in novel situations (i.e. the agent does not need to separately explore the x-reach state in the context of two different y-reach states). This aligns with the concept of \textit{disentangled representations}, characterised as disjoint representations of the underlying transformation properties of the world \cite{higgins2018definition}. That is, transformations that vary a subset of properties of the world state, while leaving all others invariant. 

In this sense, the agent can learn solely about the tool V and tool H, and when faced with a new reward location in which it requires both a positive x-reach and y-reach, it should spontaneously produce the compound tool (HV) in an optimal way. This is a simple yet non-trivial notion of innovation in which the agent does not merely just discover a new tool (as in experiment 2). The agent is able to encounter a new situation (reward location), understand the structure of the required solution (both a non-zero x-reach and y-reach) and generate the required solution (tool HV). We can think of this as a simple example of ‘one-shot’ generalisation to novel stimuli \cite{rezende2016oneshot}\cite{smith2020active}.

\begin{wrapfigure}{r}{0.5\linewidth}
\vspace{.cm}   
 \includegraphics[width=\linewidth]{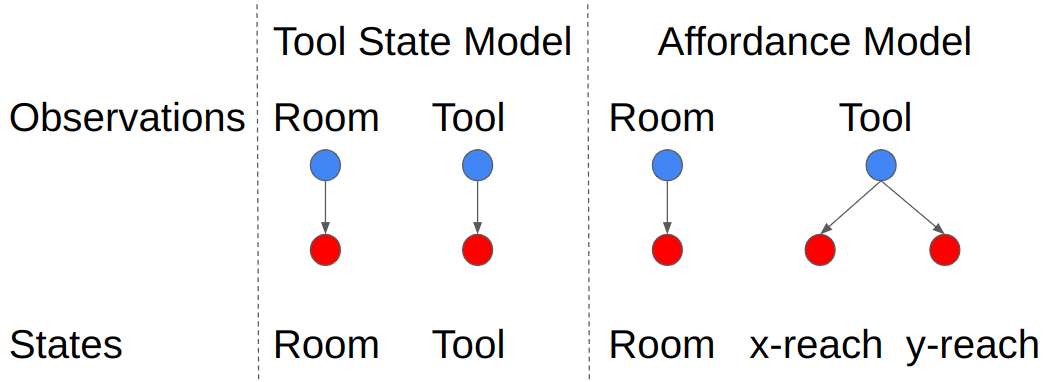}
\caption{In the Tool State Model used of experiment 2, there is a one to one mapping between the tools the agent observes, and the internal representations it has for them (None, V, H, HV). In the Affordance Model in experiment 3, the agent separates the lantent tool states into properties of x-reach and y-reach}
\label{compare}
\vspace{-0.5cm}
\end{wrapfigure} 

To test this hypothesis, we have the agent learn the entries of the transition likelihood distribution model from scratch (i.e. we initialise it as a uniform distribution as in experiment 2). However, our transition likelihood now includes the new factorised tool states (see Fig.~\ref{compare}). In a continual learning task, we present the agent with the adjacent rooms (which only require the learning about H and V) and then test it on the northeast room (which requires tool HV).

\begin{figure}
 \vspace{-0.4cm}
 \captionsetup{justification=raggedright}
\centering
\begin{subfigure}{.33\textwidth}
  \centering
  \includegraphics[width=.9\linewidth]{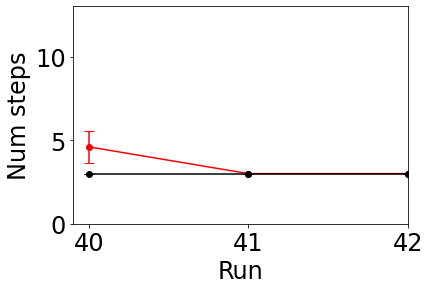}
    \caption{Reward in northeast room}
  \label{final3:steps_2}
\end{subfigure}%
\begin{subfigure}{.33\textwidth}
  \centering
  \includegraphics[width=.9\linewidth]{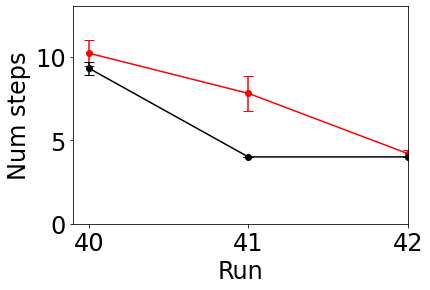}
    \caption{Reward in northwest room}
  \label{final3:steps_5}
\end{subfigure}
\begin{subfigure}{.33\textwidth}
  \centering
  \includegraphics[width=.9\linewidth]{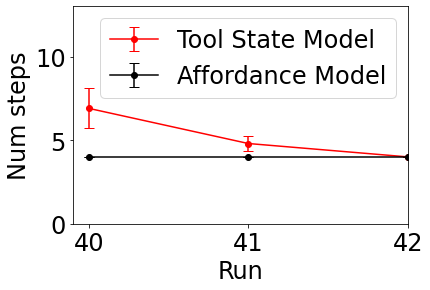}
    \caption{Northwest room, utility only}
  \label{final3:steps_5_util}
\end{subfigure}
\begin{subfigure}{.33\textwidth}
  \centering
  \includegraphics[width=.9\linewidth]{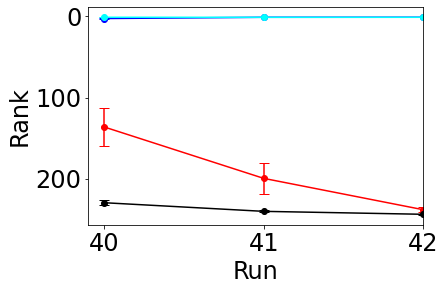}
    \caption{Reward in northeast room}
  \label{final3:ranks_2}
\end{subfigure}%
\begin{subfigure}{.33\textwidth}
  \centering
  \includegraphics[width=.9\linewidth]{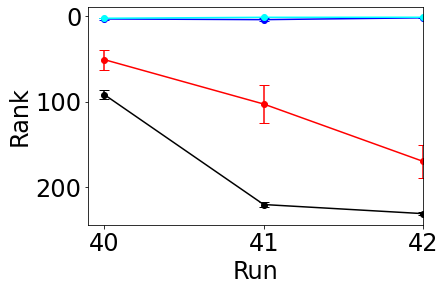}
    \caption{Reward in northwest room}
  \label{final3:ranks_5}
\end{subfigure}
\begin{subfigure}{.33\textwidth}
  \centering
  \includegraphics[width=.9\linewidth]{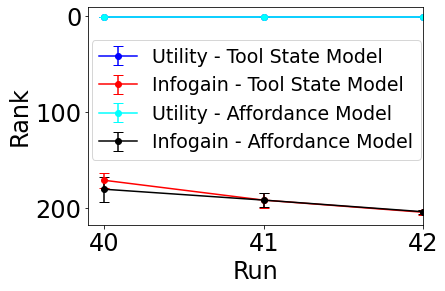}
    \caption{Northwest room, utility only}
  \label{final3:ranks_5_util}
\end{subfigure}
\caption{Factorising the hidden states into tool affordances enables the agent to perform one-shot generalisation. All graphs are the results of 3 runs following exposure to all adjacent rooms (10 runs per room) and averaged over 10 independent trials. The top panel compares the Tool State Model to the Affordance Model in terms of the mean number of steps (+/-se) taken to solve a) the northeast room b) northwest room and c) northwest room selecting policies based only on utility. The bottom panel shows the utility and information gain rank of the selected policy for d) the northeast room e) northwest room and f) northwest room selecting policies based only on utility} 
\label{final3}
\vspace{-0.4cm}
\end{figure}

Fig.~\ref{final3:steps_2} shows that, indeed, when the Affordance Model agent has only previously learned about tools H and V, it successfully creates tool HV optimally (having never seen this observation before). With the Tool State Model in experiment 2, this task was not solved optimally (as it initially took an average of roughly 5 steps to solve). As Fig.~\ref{final3:ranks_2} shows, this coincides with a greater information gain component driving action selection, meaning the agent is exploring in order to discover the compound tool. On the other hand the information gain component for the agent with the Affordance Model is much lower. This suggests that the factorisation of hidden states into affordances indeed equips the agent with the ability to leverage its current knowledge in order to compose relevant affordances and spontaneously ‘invent’ the new tool.

It is worth noting, that when repeating this experimental   procedure of exposing the Affordance Model agent to the adjacent rooms and then testing on the northwest (rather than the northeast) room, the agent does not solve this optimally, but ‘near-optimally’. As Fig.~\ref{final3:steps_5} shows, the Affordance Model agent solves this task marginally quicker than the agent with the Tool State Model, however it does not immediately find the optimal solution of 4 steps. Upon inspection of the learned transition likelihood distributions, it appears that there is a large information gain component of $\mathcal{G}$ that drives the agent to select the action ‘drop’ (and this is reflected in Fig.~\ref{final3:ranks_5}). The agent has never explored what this action ‘drop’ does in the left-hand room with no tools, and therefore it repeats this action until it no longer yields high state information gain. Once it has learned this particular fact, it then goes on to select the optimal policy and solves the task in the next 4 steps. 

To confirm that this is indeed what is causing the sub-optimal behaviour, we tailor our policy selection strategy on the critical runs. We repeat the experimental trial, but once the reward location has changed to the final northwest room, we ignore the information gain components of $\mathcal{G}$. The agent therefore selects policies based on utility alone. After this adjustment, the Affordance Model agent then solves the northwest room optimally (see Fig.~\ref{final3:steps_5_util}). Importantly, when information gain is ignored for the Tool State Model, this still does not lead the agent to solve the task optimally. This is because it does not have the required knowledge about the compound tool while the Affordance Model has all of the information it needs in order to solve the task by a process of induction.

\section{Discussion}
We have distinguished between tool use, tool discovery and tool innovation and asked what this might look like using the framework of AIF. We then ground this work with the construction of a simple model in order to take seriously this distinction and see what insights can be drawn. We provide the first evidence for the necessary properties associated with the process of tool innovation: namely that of offline induction of appropriate tool structure through composing relevant affordances.

We have identified that when solving the northwest room, the agent with the Affordance Model is not (sub-optimally) solving the task by having to discover the tool, as is the case with the agent with the Tool State Model. Rather, the agent seeks to investigate a specific state which it has never seen before and when it has sufficiently learned this fact (such that the information gain that it yields is significantly diminished), it subsequently solves the task in the optimal number of steps. Further investigation is required to ask why the utility is not enough to override this high information gain when it already has the knowledge of the correct tool to employ and the state transitions to create this tool.

We acknowledge that in our choice to factorise the hidden state of the agent’s generative model into the tool affordances of x-reach and y-reach, we play the role of an intelligent designer. Ideally, we would like to have autonomous systems that choose what to learn from the environment and factorise their model in a way that best explains the latent causes of sensory observations. Smith \textit{et al.} \cite{smith2020active} introduce an approach whereby a probabilistic generative model has flexibility in the hidden states. The idea is one of furnishing of extra “slots” in the hidden states, allowing the agent to expand its generative model to incorporate new information when encountering new concepts. A process of Bayesian model reduction then acts to prune the model, ensuring that model complexity is reduced if in fact two concepts can be explained by the same underlying cause. This approach has been further extended to deep hierarchical AIF models, facilitating the formation of flexible and generalisable abstractions during a spatial foraging task \cite{10.1371/journal.pone.0277199}.
This kind of adaptive structure learning would be useful in the context of tool innovation, allowing us to infer the best affordances to represent a tool. We therefore identify this approach as an interesting avenue for further research in the context of tool innovation in AIF agents. 

Finally, we note that our model is limited given our intentional choice to omit the sensorimotor challenges associated with both tool manipulation and tool construction. Given that tool manufacture has been identified by Beck \textit{et al.} \cite{beck2011making} as a key component in the process of tool innovation, future work should look towards constructing models which can effectively handle more physically realistic tasks.

\section{Conclusion}

Overall, we have provided a minimal description of the distinction between tool discovery and tool innovation under the formalism of active inference. We have used this to then explore a simple model of tool innovation in an AIF agent by introducing a factorisation of hidden states of the generative model into affordances. This particular structural choice affords the agent with the ability to generalise what it has learned about state transitions and conceptualise a suitable tool via a process of induction. We have discussed the implications and limitations of our results and outlined directions for further research.

\section{Author Contributions}
P.F.K. conceived the project and both designed and conducted experiments. P.C. designed and conducted experiments and wrote the manuscript. C.L.B. supervised the project.
\subsubsection*{Acknowledgements}
This research was funded under the UKRI Horizon Europe Guarantee scheme as part of the METATOOL project led by the Universidad Politécnica De Madrid.

%
\bibliographystyle{splncs04}
\bibliography{thebibliography}

\begin{thebibliography}{10}
\providecommand{\url}[1]{\texttt{#1}}
\providecommand{\urlprefix}{URL }
\providecommand{\doi}[1]{https://doi.org/#1}

\bibitem{andries2018affordance}
Andries, M., Chavez-Garcia, R.O., Chatila, R., Giusti, A., Gambardella, L.M.:
  Affordance equivalences in robotics: a formalism. Frontiers in neurorobotics
  \textbf{12}, ~26 (2018)

\bibitem{beck2011making}
Beck, S.R., Apperly, I.A., Chappell, J., Guthrie, C., Cutting, N.: Making tools
  isn’t child’s play. Cognition  \textbf{119}(2),  301--306 (2011)

\bibitem{Animaltoolusecurrentdefinitionsandanupdatedcomprehensivecatalog}
Bentley-Condit, V., Smith: Animal tool use: current definitions and an updated
  comprehensive catalog. Behaviour  \textbf{147}(2),  185 -- 32A (2010)

\bibitem{biro2013tool}
Biro, D., Haslam, M., Rutz, C.: Tool use as adaptation (2013)

\bibitem{blei2017variational}
Blei, D.M., Kucukelbir, A., McAuliffe, J.D.: Variational inference: A review
  for statisticians. Journal of the American statistical Association
  \textbf{112}(518),  859--877 (2017)

\bibitem{breyel2021beginnings}
Breyel, S., Pauen, S.: The beginnings of tool innovation in human ontogeny: How
  three-to five-year-olds solve the vertical and horizontal tube task.
  Cognitive Development  \textbf{58},  101049 (2021)

\bibitem{cabrera2020neural}
Cabrera-{\'A}lvarez, M.J., Clayton, N.S.: Neural processes underlying tool use
  in humans, macaques, and corvids. Frontiers in Psychology  \textbf{11},
  560669 (2020)

\bibitem{chappell2013development}
Chappell, J., Cutting, N., Apperly, I.A., Beck, S.R.: The development of tool
  manufacture in humans: what helps young children make innovative tools?
  Philosophical Transactions of the Royal Society B: Biological Sciences
  \textbf{368}(1630),  20120409 (2013)

\bibitem{da2022active}
Da~Costa, L., Lanillos, P., Sajid, N., Friston, K., Khan, S.: How active
  inference could help revolutionise robotics. Entropy  \textbf{24}(3), ~361
  (2022)

\bibitem{friston2012history}
Friston, K.: The history of the future of the bayesian brain. NeuroImage
  \textbf{62}(2),  1230--1233 (2012)

\bibitem{friston2017active}
Friston, K.J., Lin, M., Frith, C.D., Pezzulo, G., Hobson, J.A., Ondobaka, S.:
  Active inference, curiosity and insight. Neural computation  \textbf{29}(10),
   2633--2683 (2017)

\bibitem{gibson1977theory}
Gibson, J.J.: The theory of affordances. Hilldale, USA  \textbf{1}(2),  67--82
  (1977)

\bibitem{heins2022pymdp}
Heins, C., Millidge, B., Demekas, D., Klein, B., Friston, K., Couzin, I.,
  Tschantz, A.: pymdp: A python library for active inference in discrete state
  spaces. arXiv preprint arXiv:2201.03904  (2022)

\bibitem{higgins2018definition}
Higgins, I., Amos, D., Pfau, D., Racaniere, S., Matthey, L., Rezende, D.,
  Lerchner, A.: Towards a definition of disentangled representations (2018)

\bibitem{murphy2012machine}
Murphy, K.P.: Machine learning: a probabilistic perspective. MIT press (2012)

\bibitem{10.1371/journal.pone.0277199}
Neacsu, V., Mirza, M.B., Adams, R.A., Friston, K.J.: Structure learning
  enhances concept formation in synthetic active inference agents. PLOS ONE
  \textbf{17}(11),  1--34 (11 2022)

\bibitem{o2010innovation}
O'Brien, M.J., Shennan, S.: Innovation in cultural systems: contributions from
  evolutionary anthropology. Mit Press (2010)

\bibitem{qin2022robot}
Qin, M., Brawer, J.N., Scassellati, B.: Robot tool use: A survey. Frontiers in
  Robotics and AI  \textbf{9}, ~369 (2022)

\bibitem{reader2016animal}
Reader, S.M., Morand-Ferron, J., Flynn, E.: Animal and human innovation: novel
  problems and novel solutions (2016)

\bibitem{rezende2016oneshot}
Rezende, D.J., Mohamed, S., Danihelka, I., Gregor, K., Wierstra, D.: One-shot
  generalization in deep generative models (2016)

\bibitem{sajid2021active}
Sajid, N., Ball, P.J., Parr, T., Friston, K.J.: Active inference: demystified
  and compared. Neural computation  \textbf{33}(3),  674--712 (2021)

\bibitem{smith2020active}
Smith, R., Schwartenbeck, P., Parr, T., Friston, K.J.: An active inference
  approach to modeling structure learning: Concept learning as an example case.
  Frontiers in computational neuroscience  \textbf{14}, ~41 (2020)

\bibitem{st2008revisiting}
St~Amant, R., Horton, T.E.: Revisiting the definition of animal tool use.
  Animal Behaviour  \textbf{75}(4),  1199--1208 (2008)

\bibitem{stout2011stone}
Stout, D.: Stone toolmaking and the evolution of human culture and cognition.
  Philosophical Transactions of the Royal Society B: Biological Sciences
  \textbf{366}(1567),  1050--1059 (2011)

\bibitem{whiten2007evolution}
Whiten, A., Van~Schaik, C.P.: The evolution of animal ‘cultures’ and social
  intelligence. Philosophical Transactions of the Royal Society B: Biological
  Sciences  \textbf{362}(1480),  603--620 (2007)

\end{thebibliography}

\end{document}